\documentclass{article}

\usepackage{caption} 
\usepackage{PRIMEarxiv}

\usepackage[utf8]{inputenc} 
\usepackage[T1]{fontenc}    
\usepackage{hyperref}       
\usepackage{url}            
\usepackage{booktabs}       
\usepackage{amsfonts}       
\usepackage{nicefrac}       
\usepackage{microtype}      
\usepackage{lipsum}
\usepackage{fancyhdr}       
\usepackage{graphicx}       
\graphicspath{{media/}}     

\title{Can input reconstruction be used to directly estimate uncertainty of a regression U-Net model? - Application to proton therapy dose prediction for head and neck cancer patients 
}
\author{
  Margerie Huet-Dastarac\textsuperscript{1,*}, Dan Nguyen\textsuperscript{2}, Steve Jiang\textsuperscript{2}, John Lee\textsuperscript{1}, Ana Barragán Montero\textsuperscript{1}
}
\begin{document}
\maketitle
\begin{abstract}
Estimating the uncertainty of deep learning models in a reliable and efficient way has remained an open problem, where many different solutions have been proposed in the literature. Most common methods are based on Bayesian approximations, like Monte Carlo dropout (MCDO) or Deep ensembling (DE), but they have a high inference time (i.e. require multiple inference passes) and might not work for out-of-distribution detection (OOD) data (i.e. similar uncertainty for in-distribution (ID) and OOD). In safety critical environments, like medical applications, accurate and fast uncertainty estimation methods, able to detect OOD data, are crucial, since wrong predictions can jeopardize patients safety. \\\\
In this study, we present an alternative direct uncertainty estimation method and apply it for a regression U-Net architecture. The method consists in the addition of a branch from the bottleneck which reconstructs the input. The input reconstruction error can be used as a surrogate of the model uncertainty. For the proof-of-concept, our method is applied to proton therapy dose prediction in head and neck cancer patients. Accuracy, time-gain, and OOD detection are analyzed for our method in this particular application and compared with the popular MCDO and DE. \\\\
The input reconstruction method showed a higher Pearson correlation coefficient with the prediction error (0.620) than DE and MCDO (between 0.447 and 0.612). Moreover, our method allows an easier identification of OOD (Z-score of 34.05). It estimates the uncertainty simultaneously to the regression task, therefore requires less time or computational resources.
\end{abstract}

\section{Introduction}
Estimating the uncertainty of deep learning models in a reliable and efficient way has remained an open problem, where many different solutions have been proposed in the literature. Bayesian probability theory enables us to reason about model uncertainty with the estimation of the posterior distribution. Due to the prohibitive computational cost, most current methods are based on Bayesian approximations which aim at using the posterior predictive variance to estimate the uncertainty. The two prominent methods are deep ensembling (DE) \cite{Lakshminarayanan2017-cx} and Monte Carlo dropout (MCDO) \cite{Gal2016-bv}. DE uses the variance of the predictions of multiple networks while MCDO combines several forward passes with randomly dropped out weights. However, they require high inference time (i.e. require multiple inference passes) and MCDO has been argued to be unreliable on real world datasets \cite{Liu2021-ca}, particularly in out-of-distribution (OOD) detection (i.e. present similar uncertainty metric for in-distribution (ID) and OOD data). In safety critical environments, like medical applications, accurate and fast uncertainty estimation methods, able to detect OOD data, are crucial, since wrong predictions can jeopardize patients safety. \\

In this work, we present an alternative direct uncertainty estimation method and apply it for a regression U-Net architecture. The method consists in the addition of a branch from the bottleneck which reconstructs the input. The input reconstruction error can be used as a surrogate of the model uncertainty. Accuracy, time-gain, and OOD detection are analyzed for our method in this particular application and compared with MCDO and DE. The input reconstruction method showed a higher Pearson correlation coefficient with the prediction error than DE and MCDO. Moreover, our method allows an easier identification of OOD data, shown by a higher Z-score and no overlap between the distributions. Our method estimates the uncertainty simultaneously to the regression task, therefore requires less time or computational resources than MCDO or DE.\\

For the proof-of-concept, our method is applied to proton therapy dose prediction in head and neck cancer patients. Proton therapy is a type of radiation therapy which is a treatment modality against cancer. Planning a radiation therapy treatment is a long, complex and still mostly manual workflow as medical physicists and physicians collaborate to determine the best compromise between tumor coverage and healthy tissue sparing. Thus, the treatment plan quality depends on external and human factors, such as the available time or the clinical staff’s expertise. Automating and standardizing radiotherapy planning is therefore utterly important to ensure that all patients receive the best possible treatment, independently from the treating hospital or physician. Moreover, with the aging population and thus the increased incidence of cancer, allowing the clinicians to deliver treatments more efficiently and in closer contact with the patients will become an essential asset to maintain a high-quality standard of care.\\

Lately, the research community has progressed tremendously in the automation of the manual steps of the planning workflow through artificial intelligence (AI). Dose prediction models, for instance, predict an optimal dose trade-off for each new patient for a specific treatment modality. They can be used to guide physicians in the optimization \cite{Mashayekhi2023-mj}, as a part of automatic treatment plan generation \cite{Maes2023-tn} or for decision support in treatment indication \cite{Huet-Dastarac2023-mc}. These models are slowly getting implemented clinically \cite{De_Biase2022-zp}. The reason for the delay compared with other fields is principally due to the lack of reliable and easy to implement quality assurance tools \cite{toward}.\\ 

Direct uncertainty estimation methods try to solve this issue by proposing one shot uncertainty estimation. Methods such as direct epistemic uncertainty prediction (DEUP) \cite{Lahlou2021-zs} or layer ensembling \cite{Kushibar2022-iu} have been proposed. DEUP disentangles epistemic and aleatoric uncertainties by subtracting an estimation of the latter to the prediction of the error. The aleatoric uncertainty is approximated with the variance of labels for the same input by oracles. Despite its elegance, in applications like radiotherapy dose prediction, it would require the generation of multiple treatment plans by different experts for the same set of patients. Unfortunately, it is currently very difficult and unrealistic to obtain such data due to the time constraints of clinical environments. Layer ensembling brings a new approach to uncertainty estimation for contour segmentation by computing the variance of segmentation at different depths of the model. It has been adapted recently to dose prediction by modifying the architecture of segmenting heads by regression heads \cite{Robin_Tilman_Margerie_Huet-Dastarac_Ana_Maria_Barragan_Montero_John_Lee_undated-on}.\\

Input reconstruction has been proposed by different papers for anomaly detection by clustering the latent representation of the bottleneck of an autoencoder \cite{Legrand2019-kj}. Our proposal is to reconstruct the Computed Tomography (CT) scan, which contains the patient anatomy, given in input from the latent representation of a dose prediction model. This is done by the addition of a CT reconstruction branch from the bottleneck of the U-Net based dose prediction model (see Figure \ref{fig:arch}). The mean squared error of the CT reconstruction, masked on the body, can be used as a surrogate of the uncertainty that we evaluate by computing the correlation with the prediction loss. We also investigate its performance as an OOD detection by testing with patients that were treated for other head and neck sites (nasopharynx, larynx…). Our method successively manages to distinguish the OOD database and achieves a higher Pearson correlation coefficient between uncertainty and prediction loss than the state of the art and requires less computation time and resources as it is a direct estimation.
\section{Method}
\subsection{Architecture:} 
The architecture of our model is based on the HDUNet \cite{Nguyen2019-as}. HDUNet is a convolutional neural network, with a hybrid architecture between the popular U-Net \cite{Ronneberger2015-mw} and DenseNet \cite{Huang2017-gj}. U-Net is an encoder-decoder with skip connections that has shown outstanding performance in tasks of image-to-image mapping like segmentation mask generation. It was first published in 2015 and since then has become the state of the art for dose prediction \cite{Liu2020-fc}, \cite{BarraganMontero2019-ng}, switching from segmentation to regression. U-Net architectures can include global and local features to generate 3D voxel-wise predictions by working at different resolution levels, while dense connections (a feature of DenseNets) improve the learning by enhancing feature reuse as well as gradient backpropagation. Due to the increased memory usage of dense connections, the HD-UNet only uses them within levels of the same resolution.\\

\begin{figure}[!h]
  \centering
  \includegraphics[width=\textwidth]{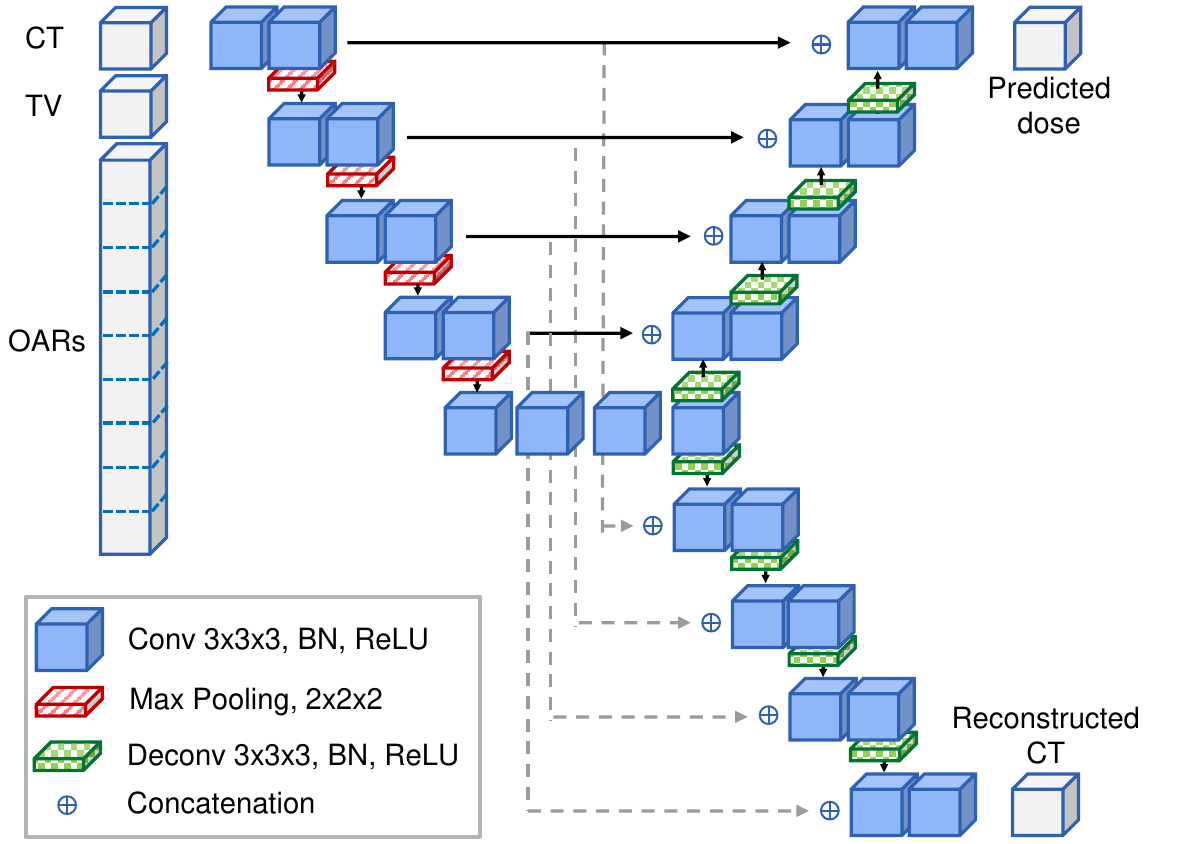}
  \caption{Architecture of the network with CT reconstruction branch}
  \label{fig:arch}
\end{figure}

Our model plugs a second decoder branch to the bottleneck after the encoder, which aims to reconstruct the CT scan given in input, like an auto-encoder would do (see Figure \ref{fig:arch}). This second decoder is composed of 4 upsampling operations with dense convolutions and skip connections stemming from the encoder. Both decoding branches, as well as the regular encoder, are trained concurrently, with the sum of two mean squared errors (MSE) as the training losses, one for dose prediction and the other for CT reconstruction. The model is fed with input data that gather the CT scan, the prescriptions of the target volumes and the binary matrices containing the volume of the OARs. 

\subsection{Data:}
\subsubsection*{Dataset 1 - ID}
The training set of the DL dose prediction network is an anonymized database of 60 patients with oropharynx cancer. The data was obtained from Cliniques Universitaires of Saint Luc (Brussels, Belgium). For each patient, a proton therapy (PT) plan was manually generated by an experienced dosimetrist. Local ethics committee approval has been obtained for the collection and use of this retrospective database. \\

The treatment planning system (TPS) used to generate the plans is RayStation 11B (Raysearch Laboratories, AB) with a research license. For consistency regarding the planning strategy and the beam configuration, all selected patients are bilateral cases and they have two dose prescription levels: 70Gy (TV\_HIGH) and 54.25Gy (TV\_LOW) on tumor and elective nodal volumes, respectively. 
PT plans are planned for pencil beam scanning delivery, using four beams at (10º, 60º), (10º, 120º), (350º, 240º) and (350º, 300º) for couch and gantry angles, respectively. The PT plans are generated using Monte Carlo dose calculation and robust worst-case optimization \cite{Souris2019-jl} on the CTV volume with 21 scenarios (2.6\% proton range error and 4mm systematic error in the three space directions). \\

Each patient has a PT dose distribution, a CT image and a set of contours for the target volumes and the OARs: brainstem, spinal cord, parotids, esophagus, oral cavity, pharyngeal constrictor muscles inferior, medium and superior, submandibular glands and supraglottic larynx.
\subsubsection*{Dataset 2 - OOD}
The second database used to investigate its performance with OOD is a set of 10 patients, obtained from the same institution. The cancer sites are also head and neck but not oropharynx: larynx, hypopharynx, nasopharynx, supraglottis and pyriform sinus. Each patient in this dataset has a CT image, the same set of contours as the ID dataset but no PT dose distribution.

\subsection{Training:}
Networks were trained for 150 epochs using the Adam optimiser to minimize the MSE loss functions from the two branches. We applied random flip as data augmentation and fed the network random patches of size 128x128x128 voxels sampled around the target volume. 
In the ID analysis as described below, we used a 11-fold nested cross-validation approach. This method entails to set aside 3 patients as an outer test set and train multiple times the networks with moving training, validation, and test sets. The validation and test sets consist of 5 patients each, and we use the remaining 47 patients for the training. The process is repeated 11 times, allowing us to obtain predictions and uncertainty values on 55 patients (number of patients per test set times number of folds). For OOD data, we selected one of the trained models for each method. All these operations were performed by a Tesla A100 GPU.

\subsection{Uncertainty estimation:}
The proposed surrogate of the uncertainty is the MSE between the clinical and reconstructed CT images. We only consider the voxels in the body contour to keep relevant information in our metric and avoid a positive bias caused by zero values of the background around the patient. Moreover, couch setup can vary and should not be considered by our metric. 

\subsection{Evaluation:}
We evaluated the performance of the input reconstruction method in three parts: its impact on the model performance, the accuracy of the uncertainty metric when fed ID data and its ability to perform OOD detection.
\subsubsection*{Impact of the additional branch on the model:}
A dose prediction model performance is generally evaluated in terms of difference of dose-volume histogram (DVH) metrics between clinical and predicted dose distributions. DVH metrics are commonly used to evaluate plans in a clinical environment. We compute the difference in DVH metrics between predictions of standard HDUNet and our model (i.e. HDUNet with input reconstruction branch). A paired-samples Wilcoxon test was performed and the threshold for statistical significance is set at p-value < 0.05. 

\subsubsection*{ID analysis:}
The dose prediction model is trained and tested on the ID dataset following the training process described in the Training subsection. At inference time, we computed the Pearson correlation coefficient between the uncertainty estimation and the dose prediction error and compared it with MCDO and DE.

\subsubsection*{OOD analysis:} 
The dose prediction model is trained on the ID dataset and the inference is performed on the ID and OOD datasets. The two distributions are displayed as histograms for CT reconstruction, MCDO and DE. We also computed the Z-score to estimate the distance between the distributions. Finally, we computed the number of patients contained in the overlap between the ID and OOD histograms for the different methods.\\

For MCDO experiments, a Bernoulli law was used to drop the weights at inference time with probabilities of 0.1, 0.2, 0.3, 0.4, and 0.5. The standard deviation is computed voxelwise on 20 inferences and then the mean over the body is considered.\\

For DE, we trained 20 models and reinitialized the weights following the He initialization \cite{He2015-im}. The standard deviation is computed voxelwise on the inference of the 20 models and then the mean over the body is considered.
\section{Results}
\begin{figure}[!h]
    \centering
    \includegraphics[width=0.8\textwidth]{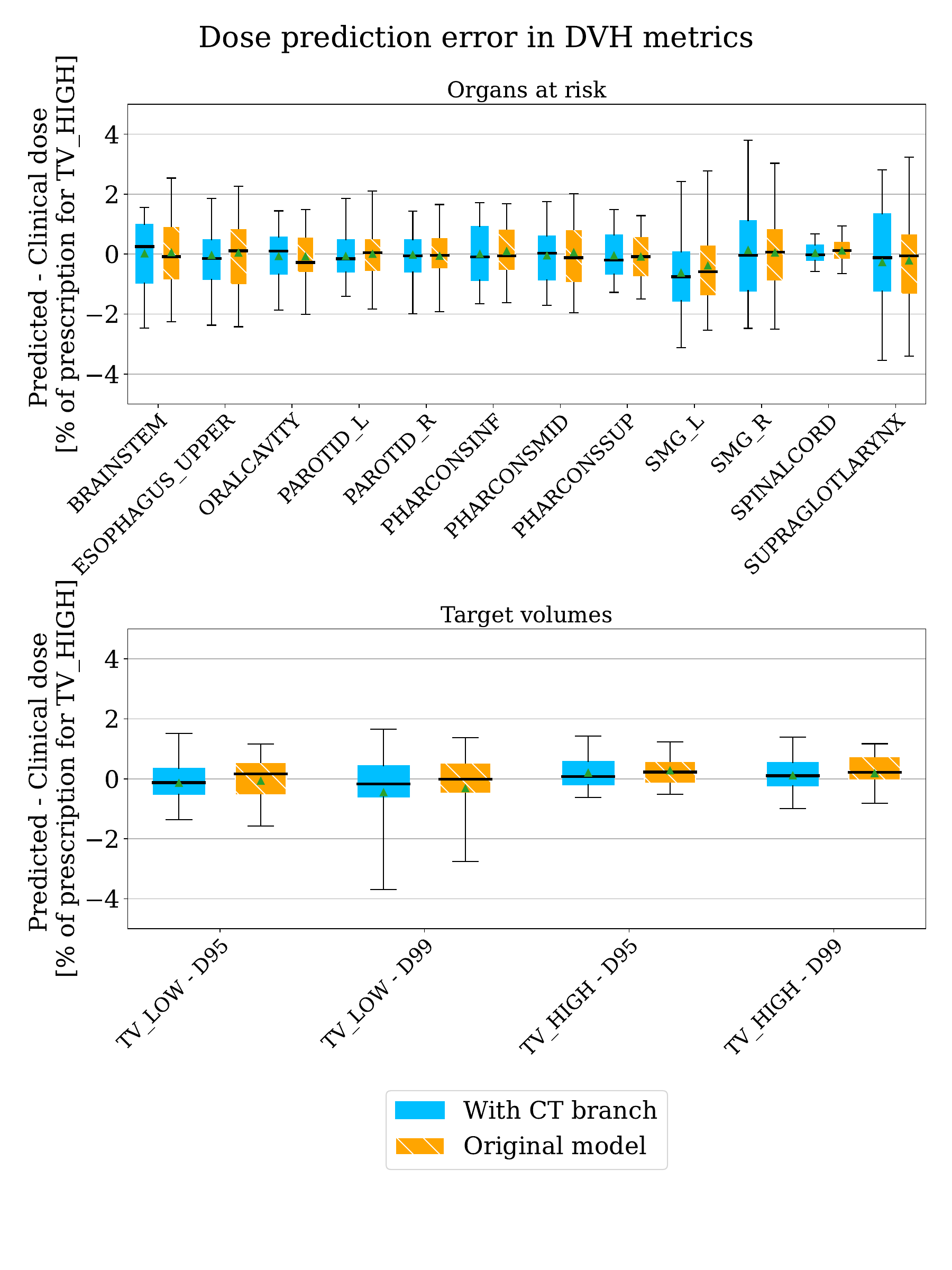}
    \caption{Performance in terms of clinical DVH metrics of the dose prediction model.}
    \label{fig:perf}
\end{figure}

The performance of the dose prediction model is shown in Figure \ref{fig:perf}. The median absolute error is close to zero in the D95 and D99 of the target volumes (Figure \ref{fig:perf}, bottom) and less than one percent of the highest prescribed dose for Dmean and D2 of OARs (Figure \ref{fig:perf}, top). The addition of the CT reconstruction branch does not impact in a statistically significant way the dose prediction of the network, as seen in Figure \ref{fig:perf} and \ref{tab:wilcoxon} for the results of the paired-samples Wilcoxon test.\\ 
{\renewcommand{\arraystretch}{2}
\begin{table}[!h]
    \centering
    \begin{tabular}{|l|c|l|c|}
    \hline
         Structure& \shortstack{\\Wilcoxon Rank\\Sum Test\\p-values} & Structure&\shortstack{\\Wilcoxon Rank\\Sum Test\\p-values} \\
         \hline
         CTV\_5425 - D95
& 0.440
 & Right parotid - Dmean
&0.712
\\
         CTV\_5425 - D99
& 0.524
 & Inferior pharyngeal constrictor muscle - Dmean
&0.540
\\
         CTV\_7000 - D95
& 0.407
 &  Middle pharyngeal constrictor muscle - Dmean
& 0.384
\\
         CTV\_7000 - D99
& 0.404
 & 
Superior pharyngeal constrictor muscle - Dmean
& 0.307
\\
         Brainstem - Dmean
& 0.388
 & Left submandibular gland - Dmean
& 0.360\\
         Esophagus upper - Dmean
& 0.535
 & Right submandibular gland - Dmean
&0.377
\\
         Oral cavity - Dmean
& 0.497
 & Spinal cord - Dmean
& 0.059\\
         Left parotid - Dmean
& 0.412
 & Supraglottic larynx - Dmean
& 0.492
\\

\hline
    \end{tabular}
    \caption{P-values paired-samples Wilcoxon Test between original model and the one with a CT reconstruction branch. Statistical significance is achieved if the p-value is inferior to 0.05. 
}
    \label{tab:wilcoxon}
\end{table}}

\begin{figure}[!h]
    \centering
    \includegraphics[width=0.8\textwidth]{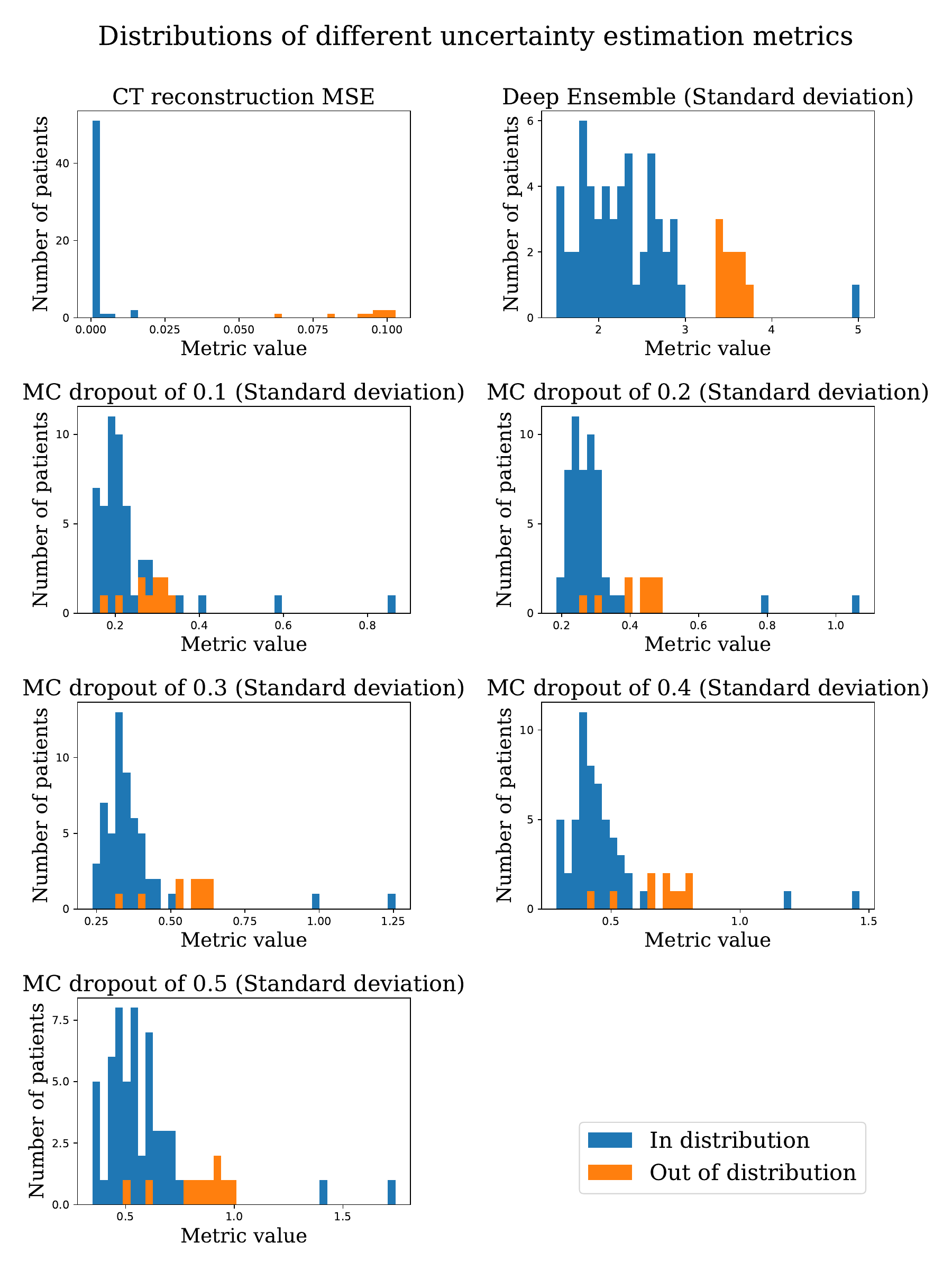}
    \caption{Distributions of the uncertainty estimation metrics}
    \label{fig:hist}
\end{figure}

The Pearson correlation coefficients between the different uncertainty metrics and the dose MSE for ID data are reported in \ref{tab:pearson}. The correlation coefficient obtained with all methods are significant as the p values are way below 0.05. The correlations are positive from 0.447 with Deep Ensemble to 0.62 with our method. For MCDO, the one with a probability 0.5 to drop yields the highest correlation of 0.612.\\
{\renewcommand{\arraystretch}{2}
\begin{table}
    \centering
    \begin{tabular}{|c|c|c|c|c|c|c|c|}
    \hline
         &  DE & MCDO (0.1)& MCDO (0.2) & MCDO (0.3) & MCDO (0.4)& MCDO (0.5)& CT reconstruction\\
         \hline
         PC&  0.447&  0.599&  0.606&  0.609&  0.613&  0.612& 0.620
\\
\hline
 p-value& 6.2e-4& 1.3e-6& 9.1e-7& 8.0e-7& 6.4e-7& 6.6e-7&4.5e-7
\\
\hline
    \end{tabular}
    \caption{Pearson correlation (PC) coefficients of mean squared error between the predicted and clinical dose in the body and mean squared error between predicted and clinical CT scan. The data was computed for in-distribution patients. DE stands for Deep Ensemble and MCDO stands for Monte Carlo dropout.}
    \label{tab:pearson}
\end{table}}
\\
In Figure \ref{fig:hist}, we can see the distribution of the different uncertainty methods for the ID and the OOD databases. While the five MCDO methods fail to distinguish OOD from ID, DE almost separates between the two distributions with only one patient on the right-hand side of the histogram. Our input reconstruction technique on the other hand allows us to easily set a threshold between ID and OOD patient data. Table \ref{tab:overlap} shows distance metrics between ID and OOD for each uncertainty estimation method.\\
{\renewcommand{\arraystretch}{2}
\begin{table}
    \centering
    
    \begin{tabular}{|c|c|c|c|c|c|c|c|}
    \hline
         & DE & MCDO (0.1)& MCDO (0.2) & MCDO (0.3)& MCDO (0.4)&  MCDO (0.5)& \shortstack{\\CT\\ reconstruction}
\\
\hline
         Z-score&  2.347&  0.368&  0.877&  1.098&  1.177&  1.124& 34.050
\\
\hline
         Intersection
& 0 & 9 & 4 & 2 & 2 & 3 & 0\\
\hline
    \end{tabular}
    \caption{Z-score and intersection (in number of patients) between in and out-of-distribution data for each method. DE stands for Deep Ensemble and MCDO stands for Monte Carlo dropout.}
    \label{tab:overlap}
\end{table}}
\\
The uncertainty estimation of our method is obtained simultaneously to the main task which takes 10 seconds. MCDO and DE, on the other hand, require 20 inference passes per patch.

\section{Discussion}
In this study, we investigate a novel architecture to directly estimate the global uncertainty and detect OOD for a regression U-Net-like architecture by adding an input reconstruction branch. We applied it to an increasingly popular application in the medical field: the prediction of radiation therapy dose distribution. The surrogate of the uncertainty is the MSE of the CT reconstructed by the second decoding branch, grafted to the regular encoder-decoder architecture of a U-Net, as illustrated in Figure \ref{fig:arch}. The addition of the CT reconstruction branch and duplicate skip connections does not impact the prediction of the dose in a statistically significant way, as seen in Figure \ref{fig:perf} and Table \ref{tab:wilcoxon}.\\ 

Table \ref{tab:pearson} shows that the CT reconstruction method presents the highest Pearson correlation coefficient compared with DE and MCDO with different dropout rates. Layer Ensembling adapted to dose prediction has shown to have slightly less good correlation between uncertainty estimation technique and MSE loss on the body contour than MCDO and DE with a difference of 3\% \cite{Robin_Tilman_Margerie_Huet-Dastarac_Ana_Maria_Barragan_Montero_John_Lee_undated-on}.  Therefore, our method is a better indicator of patients susceptible to get a prediction with larger error. \\

In Figure \ref{fig:hist}, we notice that CT reconstruction is the only method allowing a clear separation with no overlap between the ID and OOD data. This observation is confirmed with the metrics in Table \ref{tab:overlap}, where CT reconstruction has a Z-score of 34 whereas MCDO are less than 1.2 and about 2.2 for DE. Distribution shapes also differ. CT reconstruction distribution is positively skewed, DE is difficult to define and the five MCDO distributions are also positively skewed but with a wider spread. Two patients present much higher MCDO variances. They would be falsely considered as outliers if we used MCDO for OOD flagging. Despite having been used effectively for OOD detection in 2D image classification tasks in \cite{Nguyen2022-zz}, MCDO does not appear to perform well for our regression task with 3D clinical data. DE also has one patient that would have been classified as an outlier. Interestingly, the patient with highest DE or MCDO variances is the same for all metrics. After a visual check by a physician, no striking difference with other oropharynx patients was detected. The patient with the lowest CT reconstruction MSE value from the OOD dataset is a patient with a centered tumor site of supraglottis whereas the furthest apart has a tumor in the lowest part of the larynx. This observation suggests a connection between the high uncertainty and the distance between ID and OOD tumor location.\\

One of the reasons MCDO and DE sparked interest in the research community lies in how easy it is to adapt them to any model. The architecture stays identical and either several models are trained with different initialization weights like for DE or the model is inferred several times with randomly sampled weights dropped out for MCDO. However, both techniques require multiple inference passes, which means that either you need more computational resources to parallelize inferences or more time. Our solution allows us to compute the uncertainty for free, since it is computed simultaneously to the main task. This provides a huge time-gain with respect to MDCO and DE, which require on average 20-50 inference passes. In a real-time scenario such as adaptive radiotherapy workflow, reducing the time spent by a patient on the couch is crucial.\\ 

Contrary to MCDO and DE, our method outputs a single value as uncertainty measure whereas the former techniques can produce 3D uncertainty maps. Reliable uncertainty maps are definitely an interesting tool to help physicians apprehend the regions of uncertainty and would be a great addition to the interface of the clinical workflow. There are still applications where a single value can be used and where our model could be beneficial. For example, to flag OOD patients, these patients could either be redirected to another network trained on closer patients or require the attention of a clinician to proceed with the treatment. Another application could be in the curation of networks with active learning. Active learning is the task of optimizing the selection of the training set among a pool of unlabeled data \cite{Budd2021-fz,Abdar2021-eq}. It is particularly useful when the labeling task is expensive in time and resources as it is the case in treatment planning to generate the dose distribution. Current active learning methods use MCDO or DE, rank unlabeled data according to their uncertainty and select highly uncertain samples for oracles to label \cite{Budd2021-fz}. We think that active learning would also benefit from our fast method. \\

One limitation of our study is that the addition of the CT reconstruction branch has only been investigated for a regression task. This implies that both outputs (predicted dose and CT reconstruction) are trained with the same MSE loss and therefore the correlation between the error and the uncertainty is more straightforward. It would be interesting to investigate the behavior of such an addition for a task where the main loss and the reconstruction loss are different. For example, also in the field of radiation therapy, a segmentation task where Dice or cross-entropy are used as the main loss function.
\section{Conclusion}
In this study, we investigate a novel architecture to directly estimate the global uncertainty and detect out-of-distribution detection for a regression U-Net-like architecture by adding an input reconstruction branch. We applied it to the radiation therapy field where uncertainty estimation is crucial to ensure the patients safety. The task considered is dose prediction of proton therapy treatments and the reconstruction of the CT scan is the surrogate of the model uncertainty. The input reconstruction method showed a higher Pearson correlation coefficient between the dose prediction error and uncertainty estimation metric than state-of-the-art techniques like Monte Carlo dropout and Deep ensemble. Moreover, our method allows an easier identification of out-of-distribution data. Contrary to Monte Carlo dropout and Deep ensemble, the uncertainty is computed simultaneously to the main task, which is ideal in time-sensitive applications. This technique could be investigated in the future for other U-Net like architecture like segmentation or detection. 

\section*{Acknowledgments}
Margerie Huet Dastarac and Ana Barragán Montero are funded by the Walloon region in Belgium (PROTHERWAL/CHARP, grant 7289). John A. Lee is a Research Director with the F.R.S.-FNRS.

\newpage
\bibliographystyle{unsrt}  
\bibliography{bib}

\end{document}